# MNet-Sim: A Multi-layered Semantic Similarity Network to Evaluate Sentence Similarity

Manuela Nayantara Jeyaraj*[1], Dharshana Kasthurirathna[2]

[1,2]*Sri Lanka Institute of Information Technology, Malabe, Sri Lanka*

[1]nayantara_94@yahoo.com, [2]dharshana.k@sliit.lk

***Abstract*** — *Similarity is a comparative - subjective measure that varies with the domain within which it is considered. In several NLP applications such as document classification, pattern recognition, chatbot question-answering, sentiment analysis, etc., identifying an accurate similarity score for sentence pairs has become a crucial area of research. In the existing models that assess similarity, the limitation of effectively computing this similarity based on contextual comparisons, the localization due to the centering theory, and the lack of non-semantic textual comparisons have proven to be drawbacks. Hence, this paper presents a multi-layered semantic similarity network model built upon multiple similarity measures that render an overall sentence similarity score based on the principles of Network Science, neighboring weighted relational edges, and a proposed extended node similarity computation formula. The proposed multi-layered network model was evaluated and tested against established state-of-the-art models and is shown to have demonstrated better performance scores in assessing sentence similarity.*

## I. INTRODUCTION

The advent of a mass generation of digital content has brought about an inevitable overflow of user-generated content and articles across the web. However, the verifiability of all these contents is not guaranteed and keeps growing exponentially.

As such, multiple articles that exist may refer to the same subject, which can be a particular topic, an idea, an entity, or an object. Nevertheless, the information that these articles bear with regard to the subject may vary across different sources. Pieced together, the various information about a particular subject helps in building an overall perspective of that subject.

Considering the conventional ETL (Extract, Transform and Load) functions pertaining to the database context, it can be observed that data is being collected from a multitude of sources about similar subjects and is transformed into a general format where a particular subject's features are keyed in as its data values before being loaded onto the target database. Such data processing enables a system to gather more vivid and lucid information about each and every subject and store it in an easily accessible manner, paving the way for a data seeker to retrieve expected information efficiently in the future.

Hence, intense research has been focused on the area of identifying similarities between documents, sentences, and, moreover, purely facts. "Similarity" is, however, a comparative measure that varies with what is being compared and the subjective area within which it functions. As such, this research focuses on the identification of similarity within the subjective area of computational linguistics. Thereby, this research presents a novel multi-layered semantic similarity network that efficiently outperforms existing similarity inspection algorithms that evaluate sentence similarity based on a single metric by applying the principles of network science and the proposal of an extended node similarity formula.

## II. RELATED WORK

Within the domain of Natural Language Processing, Semantic Similarity has been considered a crucial aspect and paramount for many applications that lie within the related fields. Semantic Textual Similarity (STS) is a measure applied to a group of sentences or documents in order to determine their semantic similarity. This is assessed based on their overt and indirect associations or relationships with other documents or sentences in the corpus. As such, the existence of such semantic relationships is harnessed to quantify and understand the similarity between them. Semantic similarity assessments were vividly researched in [1] and [2].

Over the years, several solutions have been proposed to assess semantic similarity, and the following elucidates the current state-of-the-art models established in this particular area of NLP.

### *A. SMART*

The most general approach to semantic similarity assessment has been pre-training on massive datasets before fine-tuning on subsequent use-cases ([3]-[5]). However, such vigorous fine-tuning frequently overfit the training data close to the model when it comes to downstream tasks and thereafter struggles to generalize to new datasets. This occurs due to the insufficient amount of data available for downstream tasks and the highly complex structure of the pre-trained models.

As such, [6] proposed a model that counters the above issues and is built upon a learning framework that stabilizes the fine-tuning process of pre-trained models such that they generalize effectively to new datasets. This model is primarily built on two concepts.

- Smoothness-inducing adversarial regularization
- Bregman proximal point approximation

Smoothness-inducing adversarial regularization controls the complexity of the model during the fine-tuning process by enforcing a precise regularization technique. Having a model $f(.;\theta)$ with $n$ number of data points for the intended task represented as $(x_i, y_i)_{i=1}^{n}$ .Here, $x_i$ is the vectorized input sentence while $y_i$ bears the corresponding labels.

$$min_{\theta} F(\theta) = L(\theta) + \lambda_s R_s(\theta) \qquad (1)$$

In equation (1), $L(\theta)$ is the loss function, $\lambda_s$ is the tuning parameter and $R_s$ is the smoothness inducing adversarial regularization?

As aggressively updating the model can lead to overfitting, a Bregman proximal point approximation is applied to equation (1). This enforces a large penalty during each iteration to reduce the aggressive updates. This model has the ability to adapt well to different domains and is currently the top benchmark model for semantic textual similarity tasks.

### *B. StructBERT*

Extended from BERT (Bidirectional Encoder Representations from Transformers), which conditions on the surrounding contexts from all the layers, StructBERT takes the structure of the language into consideration in pre-training [7]. Initially, as an input text or pair of sentences are passed into the model, it converts the sentences into a single token and identifies a contextualized vector representation for that token. The vector is formed based on the words present in the text, the position in which each word occurs, and the context or the part of the text where it occurs.

The embeddings are subsequently passed on to a stack of multi-layer bidirectional transformers [8] that observe the entire sequence of the sentence or text and implement self-attention to render the textual representation. StructBERT derives concepts of pre-training objectives from BERT and extends them upon the “word structure” for single sentences and the “sentence structure” for paired sentences. Exploiting these objectives, StructBERT understands language structure in varying granularity.

### *C. T5-11B*

This model is based on transfer learning, where it is initially trained on large, data-intensive tasks, after which it is fine-tuned to specific tasks [9]. T5-11B formats any text-based task into a text-to-text problem. This follows the encoder-decoder transformer presented by [8].

First, an input token sequence is vectorized and passed onto the encoder. The encoder contains a self-attention layer and a feed-forward network that normalize the input. The input of each element is then added to the model’s output using a residual skip connection with dropout.

The decoder resembles the encoder with the exception of an additional standard attention model on top of the self-attention mechanism and the use of relative position embeddings [10] instead of fixed position embeddings. This model performs better with larger datasets. However, in most cases, these tasks have limited availability or access to such resources and can prove to be expensive.

### *D. T5-3B*

By exchanging parameters through encoder and decoder, the number of parameters is similar to BERT’s sole encoder while trying to avoid any drop in the output [9]. It also uses a masking objective to remove noise. The pre-training process is conducted on a large data corpus, and the parameters are updated during the fine-tuning process.

However, in T5-3B, the model’s capability to effectively accumulate knowledge while adding masking and then fine-tuning for a specific task is restricted by what the model understands by simply predicting sequences of corrupted text. One obstacle faced by this text-to-text technique is the risk of the model failing to produce one of the terms required during the testing process.

Working on 3 billion parameters, T5-3B is outperformed by the T5-11B model that harnesses 11 billion parameters. The T5-Large and T5-small models also follow the same architecture of the T5 family of text-to-text transformers, with a comparatively lower number of parameters as opposed to the T5-3B and T5-11B models.

The major issue with the T5 models is their size. Compared to established models such as BERT, these models are 30 times more in size and are expensive to be utilized on commodity GPU hardware systems.

### *E. Real Former*

RealFormer is a Residual Attention Layer Transformer [11]. Deriving from the conventional transformer models, RealFormer adds a residual score to each and every attention head’s raw attention score. The two attention scores are considered to render a single attention weight value using a SoftMax function.

The RealFormer outperforms the post-LN architecture as well as the Pre-LN architecture that generates vector representations of tokens with its addition of attention scores and aggregated attention weight.

Another method that had been employed to assess semantic similarity was done through Semantic similarity graphs.

### *F. Semantic Similarity Graph*

This is an unsupervised learning approach that uses pre-trained vectors to build a vector from a sentence [12],[13]. If a sentence S is made up of M number of words, then these words will be transformed into their respective word vector.

$$S = \{w_1, w_2, \ldots, w_M\}$$
$$\{w_1, w_2, \ldots, w_M\} \longrightarrow \overrightarrow{w_1}, \overrightarrow{w_2}, \ldots, \overrightarrow{w_3}\} \qquad (2)$$

Aggregating the word vectors, a sentence vector $\vec{s}$ will be obtained using the following equation.

$$\vec{s} = \frac{1}{M} \sum_{k=1}^{M} \overrightarrow{w_k} \qquad (3)$$

There are three methods to construct a similarity graph from the text: Preceding adjacent vertex (PAV), Single Similarity Vertex (SSV), and Multiple Similarity Vertex (MSV). These vary based on the method used to decide a sentence vertex's corresponding counterpart vertex [14].

***a) Preceding Adjacent Vertex*:** This graph construction method takes the general concept behind how humans read and comprehend. In trying to comprehend particular text, humans look backward for any content that they have already read [15].

Here, the similarity measure $sim(s_i, s_j)$ for a pair of sentences $s_i$ and $s_j$ will be given by;

$$sim(s_i, s_j) = \alpha\, uot(s_i, s_j) + (1 - \alpha)\, cos(\vec{s_i}, \vec{s_j}) \tag{4}$$

Where α is the balancing factor [0,1] and uot indicates the unique overlapping terms.

***b) Single Similarity Vertex:*** SSV is based on the equivalence of Semantic dependency and Semantic Similarity. Since dependency can happen in both directions, edges are allowed from following and preceding vertices. In PAV, "Precedence" and "Adjacency" are crucial constraints. But, SSV only takes the semantic similarity of sentences into consideration with directed and weighted edges where only a single outgoing edge from each vertex is allowed. An edge between two vertices will be established and assigned a weight using equation (5).

$$w(e_\{i,j\}) = \frac{cos\,(\vec{s}_i, \vec{s}_j)}{|\,\{i - j\}} \tag{5}$$

***c) Multiple Similarity Vertex:*** In contrast to PAV and SSV, which only allow a single outgoing edge from each vertex, MSV allows multiple. It chooses multiple similar sentences above a certain threshold ($\theta$) of cosine similarity [16]. Here, edges are established and assigned weights using equation (5).

***1) Computing the Similarity***

From the above-constructed graphs, the text or sentence similarity ($Sim$) will be calculated by averaging the average weight of edges departing from one vertex.

$$Sim = \frac{1}{N}\sum_{i=1}^{N}\frac{1}{L_i}\sum_{k=1}^{L_i} weight(e_{ik}) \tag{6}$$

In the above equation (6), $N$ is the Number of sentences in the text, $L_i$ is the number of outgoing edges from a particular vertex. $weight(e_{ik})$ is the corresponding weight assigned to the edge that connects nodes $i$ and $k$.

As such, the similarity between sentences in the graphs constructed using the above three methods (PAV, SSV, and MSV) can be calculated.

## III. MOTIVATION

In the former section, most of the established models are extremely large in size, expensive, and used de-noising auto encoding objectives during training. On the other hand, the semantic similarity graph methods used to measure sentence and text similarity were based on the centering theory that strictly observes the principle of the number of attention shifts in the text being inversely proportional to the coherence or the semantic similarity of that text. Consequently, these models are localized to their near-neighborhood of sentences (preceding and following sentences) and fail to efficiently portray the similarity between distant or detached sentences.

Furthermore, these approaches constrict the models from simultaneously applying different similarity metrics in evaluating the sentences that are semantically similar to the sentencing entity in the network.

Therefore, the motivation of this research is to build a model that efficiently computes the similarity between any sentence pairs, neighboring sentences as well as detached sentences allowing the semantic similarity to be simultaneously assessed based on multiple similarity measures. Hence, instead of viewing this problem on a flat plane, this research proposes a multi-layered semantic similarity network, MNet-Sim. This paper also introduces a node similarity computation formula to gauge the overall similarity between sentence pairs based on the constructed multi-layer network. Further details with regard to the proposed model are elucidated in the following section.

## IV. METHODOLOGY

In addressing the issue of effectively computing the similarity between sentences, this research introduces a multi-layered semantic network built upon different similarity measurement metrics. The advantage in using a multi-layer network with different dimensions allows the same sentences to be modeled on different planes under different variables that build relationships among the sentences. Therefore, the principle of layering the variables which need to be considered as influences in similarity computation allows for the extension or addition of more parameters [17]. As such, at present, this model takes the Cosine Similarity, Phrasal Overlap, Euclidean distance, Jaccard similarity, and Word mover's distance as dimensional parameters on a multi-layered semantic similarity graph.

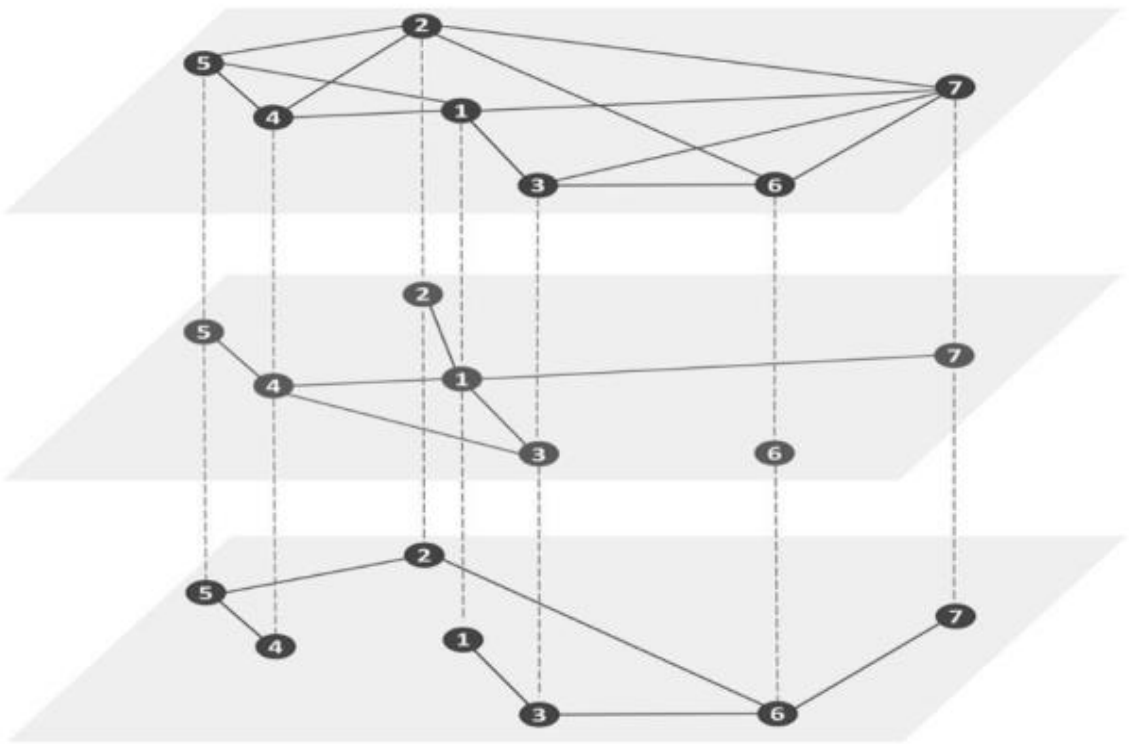

**Fig 1: Multi-layer graph with three layers or dimensions**

The conceptual layering of the model can be visualized, as shown in Fig. 1. With regard to this research, the top-most layer will correspond to the Cosine Similarity Network. The middle layer will represent the phrasal overlap network. And the bottom-most layer or any extending layers could be related to additional variables that can be introduced as an influencing factor in computing the overall similarity measure.

Each dimension in this multi-layered network will be composed of a single network of sentence vector nodes (conceptually represented by the vertices labeled 1-7). Once the individually weighted and layered network dimensions are aligned, they correspond to the same sentences, vertices, or sentence vector nodes. Subsequently, by using the proposed mathematical computation, the neighboring weighted relational edges can be aggregated, and the overall similarity between paired sentences be computed.

### A. Generating the Nodes for each Layer

Each layer is composed of nodes that represent sentence vectors. Since this research intends on identifying the similarity between sentences in passages or abstracts, using the ALBERT Sentence embedding [18] has been reported. The choice of ALBERT as the vectorization algorithm was based on its performance of having a higher data throughput, applying factorized embedding parameterization, and allowing cross-layer parameter sharing.

To generate these sentence embeddings, the data processing is done by initially tokenizing the sentences fed into the model in the form of a raw text dataset. Proceeding tokenization, the tokenized text is encoded as sentence embeddings or sentence vectors.

***a) Cosine Similarity Layer:*** the first layer of the multi-layered network has been implemented based on the cosine similarity values between sentence vector pairs. As such, the cosine similarities between all the permutations of each sentence vector are taken, and the Euclidean dot product formula is used to calculate the Cosine similarity of two non-zero vectors [19]:

$$X.Y = \|\{X\}\|\{Y\}\cos\theta \qquad (7)$$

Given two vectors $X$ and $Y$, the cosine similarity cos (θ) is represented using a dot product and a magnitude as:

$$similarity = \cos\theta = \frac{X.Y}{\|X\|\|Y\|}$$

$$\frac{X.Y}{\|X\|\|Y\|} = \frac{\sum_{i=1}^{n} X_i Y_i}{\sqrt{\sum_{i=1}^{n} X_i^2}\sqrt{\sum_{i=1}^{n} Y_i^2}} \qquad (8)$$

The computation of cosine similarities will render a vector matrix with the cosine similarities in the form of soft truth values ranging from [0,1]. Based on these cosine similarities and a decided threshold (θ), edges weighted with cosine similarities that are greater than θ will be established as relational edges between the sentence vector nodes, thus generating the cosine similarity layer in the multi-layered network.

***b) Phrasal Overlap Layer:*** The second layer of the proposed multi-layer network is built upon the Phrasal overlap scores between sentences. This measure is based on the Zipfian relationship [20]. Since conventional word overlap methods consider the sentence as a mere bag of words, the Phrasal Overlap measure has been used in order to distinguish phrases from single words in the corpus. The phrasal overlap is calculated as a nonlinear function. Here $m$ indicates the number of $i$-word phrases, and $n$ indicates the word overlaps in the sentence pairs.

$$Overlap_{phrase}(s_1, s_2) = \sum_{i=1}^{n}\sum_{m} i^2 \qquad (9)$$

To minimize the effect of outliers, equation (9) is normalized using the sum of the sentences' lengths and a tanh function as shown in equation (10).

$$Sim_o(s_1, s_2) = tanh\left(\frac{Overlap_{phrase}(s_1, s_2)}{|s_1| + |s_2|}\right) \qquad (10)$$

The phrasal overlap score, $Sim_o(s_1, s_2)$ For the sentence, pairs will be assigned as the edge weight between the nodes in the phrasal overlap layer. As such, the second layer, i.e., the phrasal overlap layer of the multi-layer network, will be constructed.

***c) Euclidean Distance Layer:*** The length of a straight line connecting two points in a Euclidean space renders the Euclidean distance between those two points. Once the points are mapped onto a Cartesian plane, this distance can be computed using the Pythagorean theorem.

Having 2 vectors represented by $x$ and $y$, $A_1$ is the frequency of vector $x$ in the first document or sentence and $A_2$ is the frequency of vector $x$ in the second document or sentence. Likewise, $B_1$ is the frequency of vector $y$ in the first document or sentence and $B_2$ is the frequency of vector $y$ in the second document or sentence.

As such, the Euclidean distance between the two vectors will be calculated as shown in equation (11).

$$d(\vec{x}, \vec{y}) = \sqrt{(A_1 - B_1)^2 + (A_2 - B_2)^2} \qquad (11)$$

The same equation can be expanded using the norm of vectors to accommodate n-dimensional vectors, as stated in equation (12).

$$d(\vec{x}, \vec{y}) = \sqrt{\sum_{i=1}^{n}(x_i - y_i)^2} \qquad (12)$$

$$Sim_e(s_1, s_2) = d(\vec{x}, \vec{y})$$

Since the Euclidean distance computes the deviation or level of difference in similarity between the vectors, the inverse of the Euclidean distance is taken as the edge weight between sentence vectors. Sentence vector pairs bearing an inverse Euclidean distance weight that is greater than the selected threshold are established as edges in the Euclidean distance dimension of the multi-layer network.

***d) Jaccard Similarity Layer:*** The Jaccard similarity, which is yet another way to compute similarity, aims to identify the shared number of tokens between sentences and divide it by the number of unique tokens found in the sentences being compared. Thereby, the common equation representing the Jaccard similarity index is shown in equation (13).

$$J(s_1, s_2) = \frac{s_1 \cap s_2}{s_1 \cup s_2}$$

$$J(s_1, s_2) = \frac{s_1 \cap s_2}{|s_1| + |s_2| - |s_1 \cap s_2|} \qquad (13)$$

$J(s_1, s_2)$, represented as $Sim_j(s_1, s_2)$ Is taken as the weight between the two sentences being compared and is used to filter out sentence pairs that bear a weight greater than the selected threshold, thus, established as edges in the Jaccard similarity dimension of the multi-layer network.

***e) Word Mover's Distance Layer:*** Word mover's distance is dependent on studying semantically meaningful representations for word tokens from local co-occurrences in sentences using embedded vectors.

Word mover's distance takes advantage of the effects of sophisticated embedding techniques such as word2vec, which produces word vectors of unparalleled consistency and generalizes well to bigger datasets. These embedding strategies show that semantic associations are often maintained when performing vector operations on word vectors.

This property of vector embeddings is used, where sentences are interpreted as a weighted accumulation of vectorized words.

Therefore, word mover's distance is the minimum total distance that words in sentence "A" need to move to precisely match the point cloud of words in sentence "B". This idea of the word mover's distance is visually represented in Fig. 2[21].

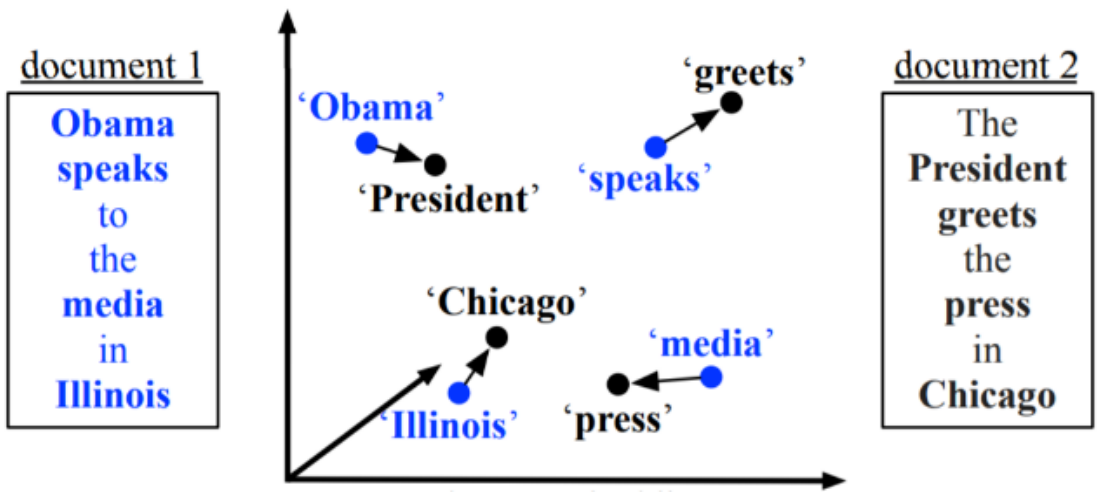

**Fig 2: Representing word mover's distance**

As such, word2vec-google-news-300 pre-trained Google news corpus word vector model with 3 billion running words was used for this research to generate the word mover distance layer.

$$\min_{T \geq 0} \sum_{i,j=1}^{n} T_{ij} c(i,j) \qquad (14)$$

$$subject\ to: \sum_{j=1}^{n} T_{ij} = d_i \quad \forall_i \{1, \ldots, i\}$$

$$\sum_{i=1}^{n} T_{ij} = d'_j \quad \forall_i \{1, \ldots, j\} \qquad (14)$$

The neighbouring nodes and edges between nodes were established based on the word mover's distance scores, computed using equation (14), between sentence pairs and represented as $Sim_w(s_1, s_2)$.

The above sections elucidate the similarity measures that have been considered in this research. Likewise, any number of layers modelled on different similarity measures can be easily added to the network in order to compute an overall similarity between sentence pairs by considering all or a selected number of measures.

### *B. Similarity Computation in the Multi-Layered Network*

To calculate the overall similarity between two sentences $s_1$ and $s_2$, i.e., $Overall_{NetSim}(s_1, s_2)$ Which is based on various similarity measures built as individual layers; initially, the local similarity between the same sentences for each layer is computed.

The following explains the steps involved in computing the Overall Network Similarity between 2 sentences, A and B, from a corpus of several sentences by considering the dimensional measure in the multi-layered network.

***a) Step 1: Computing the local layer similarity between sentence pairs***

Consider Fig. 3 as part of the cosine similarity layer representing 5 sentence vector nodes; A, B, C, D, and E, from which the similarity between sentences A and B needs to be computed.

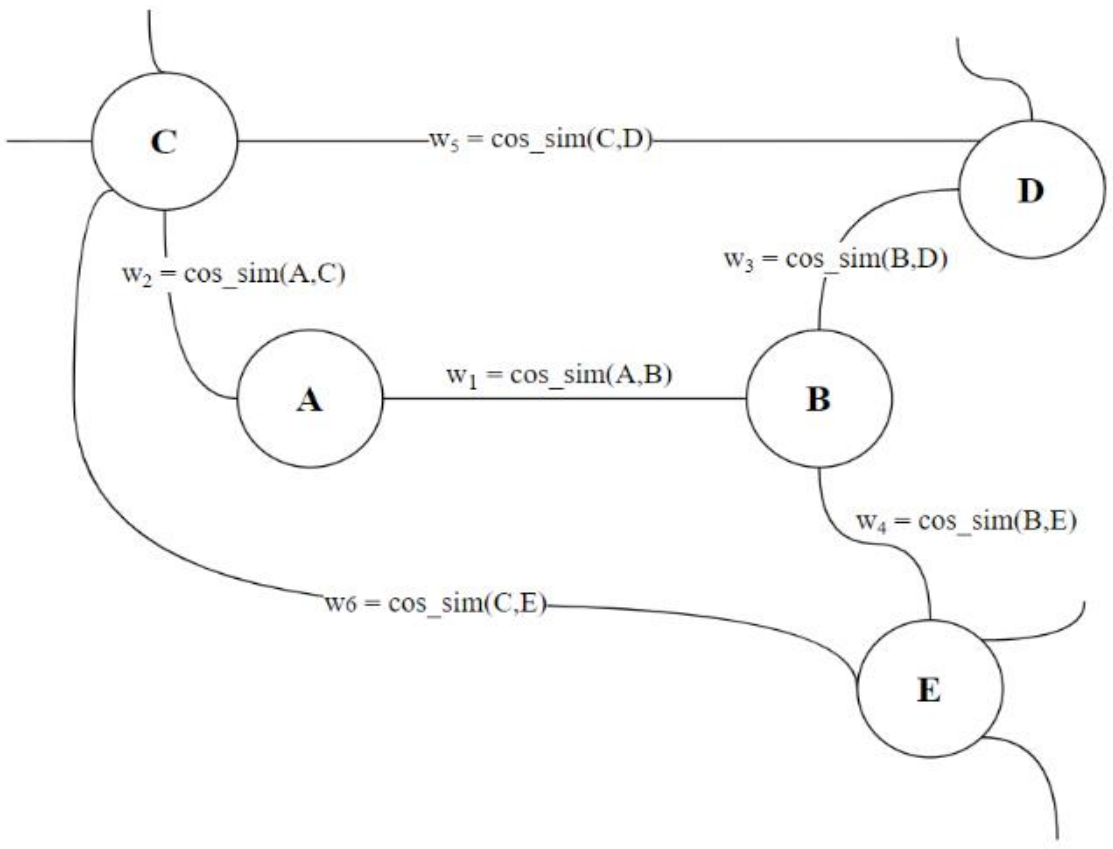

**Fig 3: Part of the Cosine Similarity Layer**

The edges between the vector nodes are established if their cosine similarity is greater than the threshold $\theta > 0.5$. And the established edges are weighted with the cosine similarity between the corresponding vector nodes.

First, the local layer similarity between sentences A and B is computed using the empirical formula proposed here below as equation (15).

$$Sim_{cosine}(A,B) = \frac{cos_{sim}(A,B) + \sum_{wtd}(A,B)}{1 + \sum_{w_adj}(A,B)} \quad (15)$$

In equation (15), $cos_{sim}(A,B)$ Refers to the weight of the edge between A & B, that is, the cosine similarity of A, B, in this case. The second term $\sum_{wtd}(A,B)$ is the weighted sum of the edges connecting the adjacent nodes of A and B. And $\sum_{w_adj}(A,B)$ is the weight of the similarity of the edges that connects the adjacent nodes of A and B.

According to Fig. 3, the adjacent node of A will be node C, and the adjacent node of B will be nodes D and E. As such, the edges that connect the adjacent nodes of A and B will be (C, D) and (C, E). Having identified the adjacent edges, the weight of the similarity of these adjacent edges, i.e., $\sum_{w_adj}(A,B)$Is calculated using equation (16).

$$\sum_{w_adj}(A,B) = weight(C,D) + weight(C,E)$$
$$where;$$
$$weight(C,D) = cos_{sim}(A,C) \times cos_{sim}(B,D),$$
$$weight(C,E) = cos_{sim}(A,C) \times cos_{sim}(B,E) \quad (16)$$

Once the weights of the similarity of these adjacent edges are calculated as shown in equation (16), the weighted sum of the edges connecting the adjacent nodes of A and B, i.e., $\sum_{wtd}(A,B)$ is computed using the following equation (17).

$$\sum_{wtd}(A,B) = weighted(C,D) + weighted(C,E)$$
$$where;$$
$$weighted(C,D) = cos_{sim}(C,D) \times weight(C,D),$$
$$weighted(C,E) = cos_{sim}(C,E) \times weight(C,E) \quad (17)$$

The first step in computing the local layer similarity between the 2 corresponding nodes A and B, in a single layer (i.e.,$Sim_{cosine}(A,B)$ in the cosine similarity layer as shown in equation (15) in the above example) will be done by substituting the values obtained using equations (16) and (17).

Similarly, moving on to the next layer in the multi-layered network, the edges and edge weights between the nodes are established based on the phrasal overlap measure, i.e., $Sim_o(A,B)$ using equation (10). Once the nodes and their edge weights are established for this network layer, the local layer similarity between sentences A and B is calculated using the same idea applied in equation (15).

Therefore, the local layer similarity of nodes A and B in the Phrasal Overlap layer in the multi-layer network, $Sim_{overlap}(A,B)$It can be calculated as shown here below in equation (18).

$$Sim_{overlap}(A,B) = \frac{Sim_o(A,B) + \sum_{wtd}(A,B)}{1 + \sum_{w_adj}(A,B)} \quad (18)$$

Here, the weight of the similarity and the weighted edges of the connections between the adjacent nodes will consider the phrasal overlap scores instead of the cosine similarity that is shown in equations (16) and (17). The same principle is applied to the other three layers and is represented as shown in equation (19)

$$Sim_{euclidean}(A,B) = \frac{Sim_e(A,B) + \sum_{wtd}(A,B)}{1 + \sum_{w_adj}(A,B)},$$
$$Sim_{jaccard}(A,B) = \frac{Sim_j(A,B) + \sum_{wtd}(A,B)}{1 + \sum_{w_adj}(A,B)}$$
$$Sim_{wordmoversdist}(A,B) = \frac{Sim_w(A,B) + \sum_{wtd}(A,B)}{1 + \sum_{w_adj}(A,B)} \quad (19)$$

In the case where more layers with different similarity measures are available in the multi-layer network architecture, the local layer similarity for that layer will be calculated in the same above manner applying that layer's corresponding similarity measure as the edge weights.

***b) Step 2: Computing the inter-layer weight***

Once each layer is established, and the respective sentences' local layer similarities are computed as suggested in step 1 above, an inter-layer weighting between the layers in the multi-layer network will be computed.

We calculate this inter-layer weight, $\nu$ , using the Pairwise Jensen-Shannon divergence measure, which normalizes the Kullback-Leibler divergence [22], taking the distribution of the individual layers.

In the case of only two layers being considered as the dimensions of the network, the inter-layer weight will be a direct computation of the Pairwise Jensen-Shannon divergence between the two layers. If more layers build up the multi-layer network, the Jensen-Shannon divergence for each paired layer permutation is computed and averaged to obtain an overall inter-layer weight.

***c) Step 3: Computing the overall sentence similarity using the aggregated layers***

Having obtained the local layer similarities of the sentences for each layer and the inter-layer weight $\nu$ using the pairwise Jensen-Shannon divergence, an overall similarity score for the sentences A and B can be computed using the following equation (20).

$$Sim(A,B) = \frac{\prod_{i=1}^{n} Sim_i(A,B)}{\nu(\sum_{i=1}^{n} Sim_i(A,B))} \quad (20)$$

Here, $i$ is any layer in the multi-layered network and hence, $Sim_i(A,B)$ corresponds to $Sim_{cosine}(A,B)$ , $Sim_{overlap}(A,B)$ , $Sim_{euclidean}(A,B)$ , $Sim_{jaccard}(A,B)$ ,

or $Sim_{wordmoversdist}(A, B)$. Applying equation (20) for a network of two layers, i.e., cosine similarity layer and phrasal overlap layer, the overall similarity between sentence pairs will be calculated as shown here below.

$$Sim(A, B) = \frac{Sim_{cosine}(A, B) \times Sim_{overlap}(A, B)}{v\left(Sim_{cosine}(A, B) + Sim_{overlap}(A, B)\right)} \quad (21)$$

If additional layers of varying similarity measures are built into the multi-layer network, their local layer similarity can be aggregated into the above equation and $Sim(A, B)$ Will render the overall similarity between any sentence pairs in the corpus using equation (20).

## V. IMPLEMENTATION SPECIFICATIONS

The proposed model was implemented according to the methodology specified in the preceding section. Thus, the specifications that dictated the implementation of this model are detailed here below.

### *A. Dataset*

For this scenario, the SemEval text corpus[1] has been used. This dataset is composed of sentences from news headlines and image captions. This dataset is built to assess the Semantic Textual Similarity (STS) based on the semantics of sentence pairs. The sentences are scored according to the golden standard between 0-5, with 0 indicating a low similarity and 5 indicating the highest similarity. Therefore, this golden-scale scoring is normalized and use as the ground truths in evaluating predicted similarity scores of the model to render its performance index metrics.

### *B. Baseline Models*

The baselines against which are considered in comparing and contrasting the proposed model are the semantic textual similarity benchmarks; SMART, StructBERT, RealFormer, T5 family of models, namely, T5-11B, T5-3B, T5-Large, and T5-Small.

### *C. Building a Multi-layer Network*

With regard to the implementation of the proposed model, the NetworkX[2] python package was utilized. This permits the creation of nodes that can bear any type of data and edges that can be weighted with arbitrary values. In this research's context, the nodes are sentence vectors, the first layer's edges are cosine similarity weighted connections, and the second layer's edges are phrasal overlap score-based connections, and so on.

Having 5 layers or similarity measures implemented on top of the sentence nodes, different networks with varying combinations of layers were modelled in order to;

- identify the number of optimal layers for the multi-layered network architecture and the proposed sentence similarity computation method.
- Identify the influence of the number of layers on the predicted similarity scores.
- Identify the metric combinations that render a higher positive correlation.
- Observe the influence of similarity metrics and distance-based metrics upon the computations via the proposed overall similarity calculation method.
- Observe the inclusion of syntactic measures on the overall similarity computation.

### *D. Performance Index Metrics*

The similarity scores computed for the baselines and the proposed model in this research are rendered as soft truth values in the range of [0,1].

As such, the following regression analysis metrics were used in analyzing the performance of the model: Mean Squared Error (MSE), Root Mean Squared Error (RMSE), Mean Absolute Error (MAE), Median Absolute Error (MedAE), Coefficient of Determination (R2), Explained Variance (EV), Maximum Error (ME). Besides these metrics, the Pearson Correlation and Spearman correlation of this model was observed against the established baseline models.

## VI. DISCUSSION

As stated in the preceding section, the proposed model was implemented as a multi-layered network with each layer corresponding to a particular similarity measure.

Out of 18-layer combinations that were implemented, 6 provided significant results and are illustrated in table 1.

Out of the 6 significant results obtained, the 2-layer network composed of the phrasal overlap and word mover's distance measures obtained the highest Pearson and Spearman correlation scores with 0.9269 and 0.9312, respectively. The three-layered network built up of the phrasal overlap, euclidean distance, and word mover's distance measures comes in at a close 0.9011 and 0.9198 Pearson and Spearman correlation, respectively.

As shown in this table, since MSE is the average squared difference between the predicted similarity score and the actual ground truth, the best possible value would be 0. As such, the experimented 2-layered network of the phrasal overlap and word mover's distance dimensions bears an optimal MSE of 0.041. The same principle applies to RMSE, which is the Standard deviation of the prediction errors. This renders the lowest possible value of 0.201 for the Multi-layered network model as the optimal performance index metric.

R2 is the coefficient of determination that observes how close the data are fitted to the regression line or decision threshold. Thereby, the first model achieves the highest R2 out of all the other combinations. Explained variance (EV) is a metric of the difference between the model's data and actual results.

[1]SemEval Dataset: https://alt.qcri.org/semeval2016/task2/
[2]NetworkX: https://networkx.github.io/

**Table 1: Evaluation Results of different combinations of layers in the multi-layer network**

| Metric | 2 Layers (po + wmd) | 3 Layers (po+ed+wmd) | 2 Layers (cs + po) | 3 Layers (cs + po + ed) | 3 Layers (cs + po + wmd) | 4 Layers (cs + po + ed + wmd) |
|---|---|---|---|---|---|---|
| ***Pearson Corr*** | 0.9269 | 0. 9011 | 0. 8684 | 0. 7854 | 0. 7421 | 0. 6821 |
| ***Spearman Corr*** | 0. 9312 | 0. 9198 | 0. 8855 | 0. 8439 | 0. 8144 | 0. 7946 |
| ***MSE*** | 0. 041 | 0. 094 | 0. 053 | 0. 095 | 0. 088 | 0. 098 |
| ***RMSE*** | 0. 201 | 0. 306 | 0. 230 | 0. 309 | 0. 296 | 0. 313 |
| ***R2*** | 0. 029 | -1. 245 | -0. 263 | -1. 281 | -1. 103 | -1. 344 |
| ***EV*** | 0. 534 | 0. 045 | 0. 409 | 0. 031 | 0. 098 | 0. 007 |
| ***ME*** | 0. 684 | 0. 979 | 0. 725 | 0. 983 | 0. 931 | 0. 996 |
| ***MAE*** | 0. 145 | 0. 232 | 0. 168 | 0. 234 | 0. 224 | 0.238 |

*cs: cosine similarity; po: phrasal overlap;*
*ed: euclidean distance; wmd: word mover's distance*

Alternatively, it is the portion of the model's overall variance that is clarified by variables that are present rather than those that occur due to error variance. Therefore, the higher the EV, the stronger the association. As such, the first network structure renders a respectable 0.534 EV that is greater than the other models. The Mean Absolute Error (MAE) is the average of absolute errors between paired observations (two sentences). With a value of 0.145, the 2-layered model attains the optimal value.

Accordingly, building a multi-layer network architecture with a minimum number of optimal layers mapped to varying similarity measures has proven to be efficient in computing the overall similarity between sentence pairs as opposed to increasing the number of layers. This observation aligns with the concept of achieving high performance with a minimum number of hidden layers in a neural network [23].

In choosing appropriate similarity measures, the introduction of a syntactic and semantic measure, rather than solely using semantic similarity measures, demonstrated enhanced performance scores.

Table 2 shows the proposed model MNet-Sim compared against the baseline models.

**Table 2: The proposed model, MNet-Sim, compared against the state-of-the-art models.**

| Model | Pearson Correlation | Spearman Correlation |
|---|---|---|
| **SMART** | 0. 929 | 0. 925 |
| **StructBERT** | 0. 928 | 0. 924 |
| **Mnet-Sim (Proposed model)** | 0. 927 | 0. 931 |
| **T5-11B** | 0. 925 | 0. 921 |
| **T5-3B** | 0. 906 | 0. 898 |
| **RealFormer** | 0. 901 | 0. 899 |
| **T5-Large** | 0. 899 | 0. 886 |
| **T5-Small** | 0. 856 | 0. 850 |

Our proposed multi-layer network model to assess semantic similarity attains a respectable Pearson Correlation of 0.927 and Spearman correlation of 0.931. Lying within the range of ±0.50 and ±1, MNet-Sim shows a strong positive correlation between the predicted similarity scores and the ground truths. As such, the proposed model is shown to have demonstrated better performance than most of the state-of-the-art models in predicting semantic similarity.

Hence, this research observes that the generation of a multi-layered network with the proposed node similarity computation method detailed in section 4.2 shows optimal performance in identifying the similarity between sentences.

This model can be applied to several Natural Language Processing applications such as plagiarism detection, information extraction, sentiment detection [24], text or document categorization, text or document summarization, topic modeling, chatbot applications, machine translation, and so on.

### *A. Limitations and Future Work*

With regard to the extension of this model, the multi-layered structure permits the inclusion of new variables as additional layers where the dimensional measures for the optimum layer combinations are application-dependent. For example, in the context of the document and its sentences need to be considered as influential factors in computing the similarity, the context can be modeled as nodes of vectors or objects with appropriately weighted edges, and the network can be further extended with more dimensions as required. However, rapidly increasing the number of layers has been demonstrated to lower the model's performance and shows that keeping the number of layers at a minimum, mapped to an optimal combination of similarity measures, is efficient as the model's complexity is controlled.

## VII. CONCLUSIONS

Semantic Similarity of text has evolved over the years from word-to-word, vector-based, and context-based assessment. This research introduces a multi-layered semantic similarity network model called MNet-Sim that efficiently evaluates the semantic similarity between sentence pairs. Here, the modeling of sentence vectors in the form of nodes in a multi-layer network allows the sentences to be simultaneously analyzed upon different similarity measures, which are then aggregated through a node similarity computation formula proposed in this paper. The proposed model was evaluated and tested against established state-of-the-art models and is shown to have demonstrated better performance scores in assessing sentence similarity.